\documentclass{article}

\usepackage{arxiv}

\usepackage[utf8]{inputenc} 
\usepackage[T1]{fontenc}    
\usepackage{hyperref}       
\usepackage{url}            
\usepackage{booktabs}       
\usepackage{amsfonts}       
\usepackage{nicefrac}       
\usepackage{microtype}      
\usepackage{lipsum}
\usepackage{graphicx}

\newcommand{\xv}{\mathbf{x}}

\newcommand{\zv}{\mathbf{z}}

\usepackage{mathtools}
\DeclarePairedDelimiter\abs{\lvert}{\rvert}%

\title{Variational auto-encoders with Student's t-prior}

\author{Najmeh Abiri\thanks{Department of Astronomy and Theoretical Physics, Lund University, Lund, Sweden} \\ najmeh@thep.lu.se \and Mattias Ohlsson\thanks{Center for Applied Intelligent Systems Research, Halmstad University, Halmstad, Sweden}     \footnotemark[1] \\mattias@thep.lu.se}

\begin{document}
\maketitle

\begin{abstract}
We propose a new structure for the variational auto-encoders (VAEs) prior, with the weakly informative multivariate Student's t-distribution. In the proposed model all distribution parameters are trained, thereby allowing for a more robust approximation of the underlying data distribution. We used Fashion-MNIST data in two experiments to compare the proposed VAEs with the standard Gaussian priors. Both experiments showed a better reconstruction of the images with VAEs using Student's t-prior distribution.    
\end{abstract}

\section{Introduction}
Variational auto-encoders (VAEs)~\cite{vae_kingma} are complex generative models that use variational inference with auto-encoder~\cite{vincent2010stacked} like architecture to learn the underlying representation. 
Unlike auto-encoders which only learn the functional representation of data, VAEs as Bayesian approaches, learn the data probability distribution, and have shown great outcome in modeling
complex distributions on increasingly larger data.

One way to improve any Bayesian model is to change the prior distribution depend on the data. Although Kingma and Welling~\cite{vae_kingma} suggested any prior from \textit{location-scale} family distribution for VAEs models, most of the studies have been done only on Gaussian priors. 
We show, with reformulating Student's t-distribution~\cite{stat_dist} and by using implicit differentiation~\cite{implicit-param}, it is possible to use the Student's t-prior. In the case of the presence of outliers in the data, the adaptive tail thickness can give enormous flexibility to the latent model.

\section{Proposed approach}
A suitable prior is essential in Bayesian analysis. In lack of knowledge of data, weakly or non-informative priors are recommended and have shown better results in many studies~\cite{gelman2008weakly}, e.g., Student's t-distribution. With knowledge, an informative distribution like the normal distribution with specified mean and covariance can be used. Although this prior has shown excellent results in many studies, O'Hagan~\cite{outlier} showed that normal priors could be outlier-resistant and never reject outliers in data modeling. Student's t-distributions allow for more heavy tails and can reject outliers when used as priors in the modeling. This motivates an implementation of the VAE with a Student's t-prior.

A location shifted p-variant Student's t-distribution has the probability density function
\begin{eqnarray}
f(z)= \frac{\Gamma(\frac{\nu +p}{2})}{\Gamma(\frac{\nu}{2})\sqrt{(\pi \nu)^p}{\sigma}}\bigg(1+ \frac{(z-\mu)^T  \sigma^{-2} (z-\mu)}{\nu}\bigg)^{-\frac{\nu+p}{2} },\nonumber
\end{eqnarray}
with $\nu$ the degree of freedom, $\mu$ the location parameter and $\sigma$ scale parameter of the distribution.
\subsection{Architecture of VAE with Student's t-distribution }
As explained in Kingma and Welling~\cite{vae_kingma}, the objective of VAE is
\begin{eqnarray}
\max_{\theta,\phi} [ \mathop{{}\mathbb{E}}_{q_\phi(\zv | \xv)} \left[\log( p_\theta(\xv | \zv))\right] - \mathbb{D_{KL}}(q_\phi(\zv | \xv)  \| p_\theta(\zv)) ],\label{equ:object}
\end{eqnarray}
with the expected likelihood term as a reconstruction loss and the Kullback-Leibler divergence term as a regularizer factor.

\begin{figure}[h!]
\centering
\includegraphics[scale=0.44]{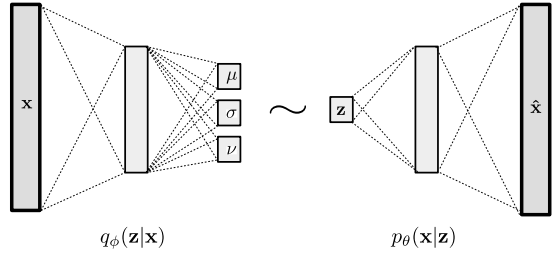}
\caption{VAE architecture with Student's t-prior. The outputs of the encoder part are the three parameters of the Student's t-distribution.}\label{Fig:vae}
\end{figure}

 The proposed structure of the VAE is illustrated in the figure \ref{Fig:vae}.
To be able to consider the Student's t-distribution as a prior for the VAE we need to adjust the objective of the network (the equation~\ref{equ:object}), to have a differentiable transformation of the probability $q_\phi(\zv | \xv)$  w.r.t. $\phi$. 

Student's t-distribution belongs to location-scale family distributions and it is possible to use the reparameterization trick,
\begin{eqnarray}
X \sim \operatorname{St}(\mu, \sigma, \nu ),\; 
X = \mu + \sigma T,\; 
T  \sim \operatorname{St}(0, I, \nu )\nonumber.
\end{eqnarray} 
Still, the replaced probability distribution is not a differentiable transformation. Another reformulation of this distribution~\cite{stat_dist} is, $x \sim \mathcal{N}(0,I)$ and $z \sim \text{Gamma}(\frac{\nu}{2},0.5)$ then $t = x/\sqrt{\frac{z}{\nu}}$ is equivalent to $t \sim \operatorname{St}(0, I, \nu )$. Figurnov and Mohamed~\cite{implicit-param} found an implicit differentiation of the cumulative function of the Gamma distribution. With the above formulation of the Student's t-distribution it is now differentiable w.r.t all three parameters.

\subsection{Training the Student's t-VAE}
In this study the prior over the latent variables is a multivariate Student's t-distribution, $p_\theta(\zv)= \operatorname{St}(0, I, \nu )$ and the variational approximate posterior is $q_\phi(\zv | \xv) = \operatorname{St}(\mu, \sigma, \nu)$. The parameters; $\mu$, $\sigma$ and $\nu$ are neural networks, as in figure \ref{Fig:vae}, with the weights $\phi$. The KL divergence of the objective takes the form
\begin{eqnarray}
&&\mathbb{D_{KL}}(q_\phi(\zv | \xv)  \| p_\theta(\zv)) =\int q_\phi(\zv | \xv) \log \left(  \frac{q_\phi(\zv | \xv) }{p_\theta(\zv)} \right)d\zv \nonumber\\
&=&-\log \lvert \sigma \rvert -\left(\frac{\nu+p}{2}\right) \mathop{{}\mathbb{E}}_{q_\phi}\left[ \log\left(1+\frac{(\zv-\mu)^T  \sigma^{-2} (\zv-\mu)}{\nu} \right)\right] \nonumber\\
&&\hspace{1.3cm} +\left(\frac{\nu+p}{2}\right)\mathop{{}\mathbb{E}}_{q_\phi}\bigg[ \log\left(1+\frac{z^2}{\nu} \right)\bigg].\label{equ:kl-d}
\end{eqnarray}

\noindent Zografos~\cite{zografos-entropy} shows that the first expectation in equation~\ref{equ:kl-d} is equivalent to
%
%
%
\begin{eqnarray}
\psi\left(\frac{\nu+p}{2}\right)-\psi\left(\frac{\nu}{2}\right)\nonumber,
\end{eqnarray}
where $\psi(x)=\frac{{\Gamma^\prime(x)}}{\Gamma(x)}$ is the digamma function. The second expectation in equation~\ref{equ:kl-d} is
\begin{eqnarray}
\frac{\Gamma(\frac{\nu+p}{2})}{\Gamma(\frac{\nu}{2})\sqrt{(\pi \nu)^p}\abs{\sigma}} \int \bigg[ 1+\frac{(\zv-\mu)^T \sigma^{-2} (\zv-\mu)}{\nu}  \bigg]^{\frac{-\nu+p}{2}} \log\left(1+\frac{\zv^2}{\nu} \right)dz\nonumber,
\end{eqnarray}
that requires numerical integration. With the expression of the KL-divergence and a suitable reconstruction error, taking the type of the data into account, stochastic gradient methods can now be used to train the models. 

\section{Experiments}
We trained VAE with two different priors, a multivariate Gaussian (VAE-G) and our proposed multivariate Student's t-prior (VAE-St) on the Fashion-MNIST~\cite{fashion} dataset. Although Fashion-MNIST has the same shape and grayscale format as MNIST, the images are more diverse which force the models to learn more advanced features. In each experiment, the models are trained on $60000$ images and tested on $10000$. During training, a fixed validation dataset (20\% of training data) was used for model selection. To estimate the maximum likelihood $\mathop{{}\mathbb{E}} \left[\log( p_\theta(\xv | \zv))\right]$, we used cross-entropy. In all experiments, computations were performed on a single GPU (GeForce GTX 1080 Ti). 

\subsection{VAEs with complete data}
Although the data are images, both VAE-G and VAE-St used multi-layer perceptrons (MLP) for both the encoder and the decoder. The choice of MLPs instead of convolutional neural networks was to simplify the hyperparameter optimization process. To evaluate the performance of VAEs, the generated images were compared with the original ones using the structural similarity~\cite{ssim} (SSIM) index. Unlike mean squared difference that resembles the images on individual pixels, SSIM treats images more holistically by comparing local regions of the images. The best models, after extensive random search~\cite{bergstra2012random} on hyperparameters, are shown in table \ref{Tab:hyper}.

\begin{figure}[h!]
\centering
\includegraphics[scale=0.35]{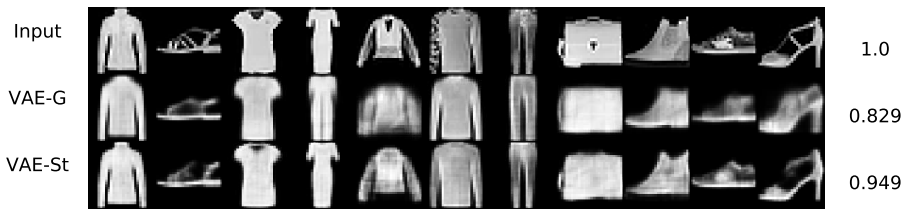}
\caption{Generated images from the two VAEs. The first row shows input images from the test set, row two and three are the corresponding conditionally generated images from VAE-G and VAE-St, respectively. Note, only one sampled image for each input is shown. The last column shows the average SSIM index over the test data.}\label{Fig:no_mask}
\end{figure}

\begin{table}[h!]
  \centering
  \begin{tabular}{|c|c|c|c|c|c|c|c|}
    \hline
    Model & encoder & decoder & act  &lr& batch & epochs&  SSIM\\
    \hline
    VAE-G & $[1453, 44]$ &  $[44, 702]$ & $\tanh$ &$10^{-3}$ &  $500$ & $200$&  $0.829$ \\
    \hline
    VAE-St &  $[1419, 42]$ & $[42, 759]$& relu &  $10^{-4}$  & $500$ & $150$&  $\mathbf{0.949}$  \\
    \hline
  \end{tabular}
  \caption{The best sets of hyperparameters for each model, together with the average SSIM index over the test data. The numbers for the encoder and the decoder are the size of the hidden layer and the dimension of the latent space (last number in the encoder list and first number in the decoder list).}\label{Tab:hyper}
\end{table}
Examples of images generated by the two VAEs are shown in figure \ref{Fig:no_mask}. In the first row, there are $11$ randomly selected images, at least one from each class, from Fashion-MNIST test set. The second and third rows are the corresponding outputs of VAE-G and VAE-St respectively.

\subsection{VAEs with incomplete data}
\begin{figure}[h!]
\centering
\includegraphics[scale=0.35]{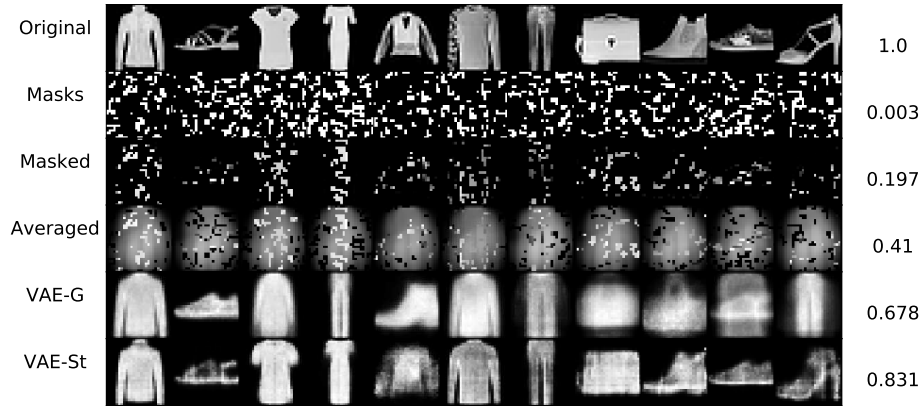}
\caption{The first row shows examples of original test images. Applying the masks, shown in row two, results in the corrupted images in row three. The input of the trained VAE-G and VAE-St is the fourth row (average over known pixels). The fifth and sixth rows show one sampled image for each input of VAE-G and VAE-St, respectively. The last column shows the average SSIM index over the test data.}\label{Fig:t-dist}
\end{figure}

The importance of using Student's t-distribution as a prior in VAE is even more evident in the presence of missing data. The Student's t-prior, with its adjustable tails, lets the VAEs find more profound features which improve the estimation of missing data. To evaluate how well VAEs can impute missing data, we manually corrupted Fashion-MNIST which enables a comparison with original images. Image data have high local correlations; therefore a random pixel-wise corruption will not be difficult enough. Instead, we used  Square-lattice Ising-like~\cite{ising}, $28\times28$ images of masks. In such a mask approximately $78\%$ of the pixels are corrupted, i.e., replaced by NaN. The second row of figure \ref{Fig:t-dist} shows examples of such masks.

To train the VAE with missing data, we replaced each NaN pixel with the average over the known samples for that pixel. The loss function is calculated only on non-missing pixels. Similar to the previous experiment, the VAEs are evaluated with the SSIM index. The best models, after hyperparameter search, are shown in table \ref{Tab:hyper_imput}. 
\begin{table}[h!]
  \centering
  \begin{tabular}{|c|c|c|c|c|c|c|c|}
    \hline
    Model & encoder & decoder & act  &lr& batch & epochs&  SSIM\\
    \hline
    VAE-G & $  [905, 24] $ &  $ [24, 902] $ & relu &$10^{-3}$ & $500$  &$300$ &  $0.678$ \\
    \hline
    VAE-St&  $[1139, 463, 37]$ &  $[37, 1308]$ & $\tanh$ &$10^{-3}$ &  $4000$ & $300$&  $\mathbf{0.831}$ \\
    \hline
  \end{tabular}
  \caption{The best sets of hyperparameters for each model, together with the average SSIM index over the test data. The numbers for the encoder and the decoder are the size of the hidden layers and the dimension of the latent space (last number in the encoder list and first number in the decoder list).}
  \label{Tab:hyper_imput}
\end{table}

Figure~\ref{Fig:t-dist} shows example images from the test set. The first and the third row show the original (uncorrupted) images and the images after applying the Ising masks, respectively. The fourth row shows the actual input images to the VAEs where missing pixels are replaced by mean values over known data (mean imputation). Mean imputation does not work for this degree and structure of missing data. Both VAE-G and VAE-St improves over mean imputation as indicated by the SSIM index. From a visual inspection and as shown by the SSIM index, the use of a Student's t-prior is improving the reconstructed images.

\section{Conclusion}
In this paper, we have implemented a variational auto-encoder with a multivariate Student's t-prior distribution which was accomplished using the implicit reparameterization gradient of the Gamma function~\cite{implicit-param}.
Two different experiments were performed on Fashion-MNIST data with both VAE-G and VAE-St. In the first experiment, we trained both models with complete Fashion-MNIST images and evaluated the models on test data. For the second experiment, we used VAE-G and VAE-St as imputation methods to recover missing pixels on corrupted Fashion-MNIST data. Both models used cross-entropy as loss function and their corresponding KL divergence as regularization. 
With the change of prior from Gaussian to Student's t-distribution in the VAE, we were able to get less blurry and more detailed images for both experiments that were quantified by the SSIM index being larger for VAE-St in both trials. Although the computations for the VAE with Student's t-prior distribution are more extensive, the difference in execution time on a GPU was not significant. We believe that using the Student's t-prior distribution for the VAE allows modeling of more complex data, such as medical data with a mixture of different input variable types. A typical challenge in working with medical data is to impute missing values. A flexible generative model such as the proposed VAE-St can have a significant impact on imputation, which will be our focus for future developments.

%
%
\bibliographystyle{unsrt}
\bibliography{vae-st}

\begin{thebibliography}{10}

\bibitem{vae_kingma}
Diederik~P Kingma and Max Welling.
\newblock Auto-encoding variational bayes.
\newblock {\em arXiv preprint arXiv:1312.6114}, 2013.

\bibitem{vincent2010stacked}
Pascal Vincent, Hugo Larochelle, Isabelle Lajoie, Yoshua Bengio, and
  Pierre-Antoine Manzagol.
\newblock Stacked denoising autoencoders: Learning useful representations in a
  deep network with a local denoising criterion.
\newblock {\em Journal of Machine Learning Research}, 11(Dec):3371--3408, 2010.

\bibitem{stat_dist}
Catherine Forbes, Merran Evans, Nicholas Hastings, and Brian Peacock.
\newblock {\em Statistical distributions}.
\newblock John Wiley \& Sons, 2011.

\bibitem{implicit-param}
Michael Figurnov, Shakir Mohamed, and Andriy Mnih.
\newblock Implicit reparameterization gradients.
\newblock {\em arXiv preprint arXiv:1805.08498}, 2018.

\bibitem{gelman2008weakly}
Andrew Gelman, Aleks Jakulin, Maria~Grazia Pittau, Yu-Sung Su, et~al.
\newblock A weakly informative default prior distribution for logistic and
  other regression models.
\newblock {\em The Annals of Applied Statistics}, 2(4):1360--1383, 2008.

\bibitem{outlier}
Anthony O'Hagan.
\newblock On outlier rejection phenomena in bayes inference.
\newblock {\em Journal of the Royal Statistical Society. Series B
  (Methodological)}, pages 358--367, 1979.

\bibitem{zografos-entropy}
Konstantinos Zografos et~al.
\newblock On maximum entropy characterization of pearson's type ii and vii
  multivariate distributions.
\newblock {\em Journal of Multivariate Analysis}, 71(1):67--75, 1999.

\bibitem{fashion}
Han Xiao, Kashif Rasul, and Roland Vollgraf.
\newblock Fashion-mnist: a novel image dataset for benchmarking machine
  learning algorithms.
\newblock {\em arXiv preprint arXiv:1708.07747}, 2017.

\bibitem{ssim}
Zhou Wang, Alan~C Bovik, Hamid~R Sheikh, and Eero~P Simoncelli.
\newblock Image quality assessment: from error visibility to structural
  similarity.
\newblock {\em IEEE transactions on image processing}, 13(4):600--612, 2004.

\bibitem{bergstra2012random}
James Bergstra and Yoshua Bengio.
\newblock Random search for hyper-parameter optimization.
\newblock {\em Journal of Machine Learning Research}, 13(Feb):281--305, 2012.

\bibitem{ising}
Rodney~J Baxter.
\newblock {\em Exactly solved models in statistical mechanics}.
\newblock Elsevier, 2016.

\end{thebibliography}


\end{document}